\documentclass[conference]{IEEEtran}
\IEEEoverridecommandlockouts
\usepackage[numbers, sort]{natbib}
\usepackage{amsmath,amssymb,amsfonts}
\usepackage{algorithmic}
\usepackage{graphicx}
\usepackage{textcomp}
\usepackage{xcolor}
\usepackage[capitalise]{cleveref}
\usepackage[acronym]{glossaries}
\usepackage{siunitx}
\usepackage{caption}
\usepackage{subcaption}

\DeclareSIUnit{\belmilliwatt}{Bm}
\DeclareSIUnit{\dBm}{\deci\belmilliwatt}

\def\BibTeX{{\rm B\kern-.05em{\sc i\kern-.025em b}\kern-.08em
    T\kern-.1667em\lower.7ex\hbox{E}\kern-.125emX}}

\newacronym{aip}{AiP}{Antenna in Package}
\newacronym{ros}{ROS}{Robot Operating System}
\newacronym{iac}{IAC}{Indy Autonomous Challenge}
\newacronym{cw}{CW}{Continuous Wave}
\newacronym{fmcw}{FMCW}{Frequency Modulated Continuous Wave}
\newacronym{fft}{FFT}{Fast Fourier Transform}
\newacronym{aoa}{AoA}{Angle of Arrival}
\newacronym{fov}{FoV}{Field of View}
\usepackage{eso-pic}
\usepackage{url}
\AddToShipoutPictureBG*{
  \AtPageUpperLeft{%
    \put(0,-40){\raisebox{15pt}{\makebox[\paperwidth]{\begin{minipage}{21cm}\centering
      \textcolor{gray}{This article has been accepted for publication in the proceedings of 2023 IEEE International Workshop on Advances in Sensors and Interfaces. Content may change prior to final publication. DOI: 10.1109/IWASI58316.2023.10164312} 
    \end{minipage}}}}%
  }
  \AtPageLowerLeft{%
    \raisebox{25pt}{\makebox[\paperwidth]{\begin{minipage}{21cm}\centering
      \textcolor{gray}{ \copyright 2023 IEEE.  Personal use of this material is permitted.  Permission from IEEE must be obtained for all other uses, in any current or future media, including reprinting/republishing this material for advertising or promotional purposes, creating new collective works, for resale or redistribution to servers or lists, or reuse of any copyrighted component of this work in other works.
      }
    \end{minipage}}}%
  }
}
\begin{document}

\title{Towards Robust Velocity and Position Estimation of Opponents for Autonomous Racing Using Low-Power Radar}

\author{\IEEEauthorblockN{Andrea Ronco, Nicolas Baumann, Marco Giordano, Michele Magno}
\IEEEauthorblockA{\textit{Center for Project Based Learning - D-ITET} \\
\textit{ETH Zurich}\\
Zurich, Switzerland \\
\{name.surname\}@pbl.ee.ethz.ch}
}
\IEEEoverridecommandlockouts
\maketitle
\IEEEpubidadjcol

\begin{abstract}
This paper presents the design and development of an intelligent subsystem that includes a novel low-power radar sensor integrated into an autonomous racing perception pipeline to robustly estimate the position and velocity of dynamic obstacles. The proposed system, based on the Infineon BGT60TR13D radar, is evaluated in a real-world scenario with scaled race cars. The paper explores the benefits and limitations of using such a sensor subsystem and draws conclusions based on field-collected data.
The results demonstrate a tracking error up to 0.21~$\boldsymbol{\pm}$~0.29 \si[detect-all=true]{\metre} in distance estimation and 0.39~$\boldsymbol{\pm}$~0.19 \si[detect-all=true]{\metre / \second} in velocity estimation, despite the power consumption in the range of 10s of milliwatts.
The presented system provides complementary information to other sensors such as LiDAR and camera, and can be used in a wide range of applications beyond autonomous racing.
\end{abstract}


\section{Introduction}\label{sec:intro}
The field of autonomous racing has gained significant attention in recent years, with the development of self-driving cars and the increasing popularity of motorsports, which opens the opportunity to enable knowledge transfer from academia to industry \cite{catalyst0, ar_survey, tcdriver, map}.
A fundamental component of autonomous systems is perception, namely the part of the system that enables the vehicle to observe and acquire information on the surrounding environment, which is necessary to generate appropriate responses  \cite{dronerace_jfr, okelly2020f1tenth, indyautonomous, roborace}.

LiDARs, cameras, and radars are the main exteroceptive sensors used for perception purposes on autonomous vehicles \cite{8585340}.
Cameras use visible light to capture images and record video. They can provide high-resolution images and color information, but they rely on good lighting conditions and can be affected by shadows and reflections.


LiDARs use infrared lasers to create three-dimensional point clouds of the surrounding environment, measuring the time of flight of the laser beam from the sensor to the reflecting objects. Due to the rotating parts, LiDARs have a fairly high cost and large size. They only provide ranging information and they are severely affected by environmental conditions (rain, snowflakes, and fog), as well as by the reflectivity properties of the targets. For example, opaque black objects are often difficult to detect with this technology \cite{hokuyo}.
Radars use radio waves with various frequencies and modulations to detect objects in their field of view. Thanks to the larger radio wavelengths, radars are less affected by adverse weather conditions, and they can measure reliably through raindrops, snowflakes, and dust \cite{hydrafusion, hrfuser}.
Radars on the other hand are very robust against the aforementioned object reflectivity and further robust to adverse weather conditions, such as rain and snowflakes \cite{badweather}. 
Depending on the technology they can also provide valuable relative velocity information and spatial position of objects, which is highly beneficial in racing contexts \cite{ar_survey, ayoubMotionPredictor}.
Each one of these technologies has its own strengths and weaknesses, and they are often used in combination to provide a more comprehensive view of the environment. 

Emerging novel low-power radars are capable of high-resolution measurements with a peak power consumption in the order of 100s of milliwatts, and an average consumption down to less than \SI{10}{\milli\watt} \cite{scherer2021tinyradarnn}. Moreover, their low cost and reduced form factor, also thanks to the \gls{aip}, make them an attractive option for achieving robust velocity and position estimation of opponents in the context of autonomous driving \cite{saponara2019radar} and racing, especially on small-scale vehicles. 
LiDAR and radar sensor fusion is a popular technique in autonomous driving, robotics, and other applications where accurate and reliable sensing is critical \cite{badweather, hrfuser, hydrafusion}. By combining the strengths of both technologies, it is possible to create a more comprehensive view of the environment, allowing for safer, more efficient, and more robust operations \cite{hrfuser, hydrafusion}.

Autonomous racing is gaining popularity in the research community as an application for novel perception and control algorithms \cite{ar_survey, indyautonomous, okelly2020f1tenth, amz_fullstack}. Especially autonomous driving and racing on small-scale vehicles have been used to develop and test algorithms for autonomous vehicles in a safe and efficient way \cite{liniger_mpcc, okelly2020f1tenth}. 
One example of an autonomous vehicle project is the F1TENTH project \cite{okelly2020f1tenth, F1TENTHbuild}, which is a global engineering competition for university students that challenges teams to design, build, and race a single-seat racing car. The F1TENTH vehicle consists of various sensor modalities such as LiDAR, camera, radar, and optical flow to achieve a high level of autonomy in racing scenarios. The perception system of the F1TENTH vehicle is crucial for achieving a good racing performance \cite{ar_survey} by accurately detecting opponents, estimating their positions and velocities, and avoiding collisions.

In this context, the use of radar solutions for opponent detection and velocity estimation can be a valuable addition to the existing sensor suite of the F1TENTH vehicle.
The adoption of low-power, small radar options enables the integration of the system in the vehicle without affecting the overall weight and aerodynamics and paves the road for a wider adoption on small-scale vehicles by assessing its effectiveness in a real racing scenario.

This paper presents a preliminary evaluation of a small-size, low-power radar sensor that operates in the \SI{60}{\giga\hertz} frequency band to asses its performance in autonomous vehicles and perception applications.
While more established radar systems already exist for these tasks, we argue that the advantages of this small-scale, integrated solution could enable new classes of vehicles to take advantage of radar technology.
We evaluate the sensor for range and velocity estimation with different targets and set up the grounds for future work involving sensor fusion of LiDAR and radar data on our racing platform.
 
The rest of the paper is organized as follows: In \cref{sec:rel_work} we present existing work on the topic of autonomous driving and radar technology for perception and autonomous vehicles, and we summarize the contributions of the paper. In \cref{sec:radar} we introduce a taxonomy of the most common radar technologies and provide the required theoretical background on radars. The evaluation setup is described in \cref{sec:setup}, while the results of our analysis are reported in \cref{sec:results}. We draw conclusions and present future work in \cref{sec:conclusion}.   

\section{Related Work} \label{sec:rel_work}
Radars are commonly used in autonomous driving applications to provide information about the surrounding environment, including the position and velocity of other vehicles on the road. However, accurately estimating the position and velocity of opponents in autonomous racing can be challenging due to factors such as noisy data, complex dynamics of the opponents, and high speeds.

Radars showed promising results when deployed on cars, both for driving assistance devices and autonomous driving. Previous work showed how radar, especially when fused with other kinds of sensors, can reliably detect road boundaries and other vehicles \cite{nabati2020radar}, especially with adverse weather conditions \cite{wang2018vehicle}.
Previous work already showed how \textit{mmWave} radars can be used to estimate the velocity and position of other vehicles in autonomous driving scenarios\cite{zhou2020mmw}. However, the radars used were automotive-grade radars, which are orders of magnitude more expensive and power-hungry than the novel \textit{mmWave} radars used in our work.

Radars have also been used in autonomous racing, as reported in \cite{wischnewski2022indy, betz2022tum}, where the authors argue their use in the \gls{iac}, a competition of fully autonomous race cars at the Indianapolis Motor Speedway, promoting innovation and technological advancements in autonomous vehicle technology \cite{indyautonomous}.

On the notes of the \gls{iac}, the F1TENTH association is a student competition of 1:10 scaled race cars, with the similar goal of competing on miniature racetracks for \textit{time-trials} and \textit{head-to-head} races. Due to the compact dimension of the car, it is not possible to use automotive radars on this platform, and therefore opponent estimation has traditionally been carried out through LiDAR and/or camera-based sensing modalities.

This paper wants to investigate novel low-power radars in a tiny form factor in autonomous vehicles' perception, in particular regarding other vehicles' detection, ranging, and exteroceptive velocity estimates.

The contributions of this paper are as follows:
\begin{itemize}
    \item A novel low-power radar sensor is introduced and integrated into an autonomous racing perception pipeline.
    \item The sensor is evaluated in a real-world scenario with scaled race cars.
    \item Benefits and limitations of such a sensor are explored and conclusions are drawn on the base of field-collected data.
\end{itemize}
\section{Radar Background}\label{sec:radar}

\begin{figure}
    \centering
    \includegraphics[width=\linewidth]{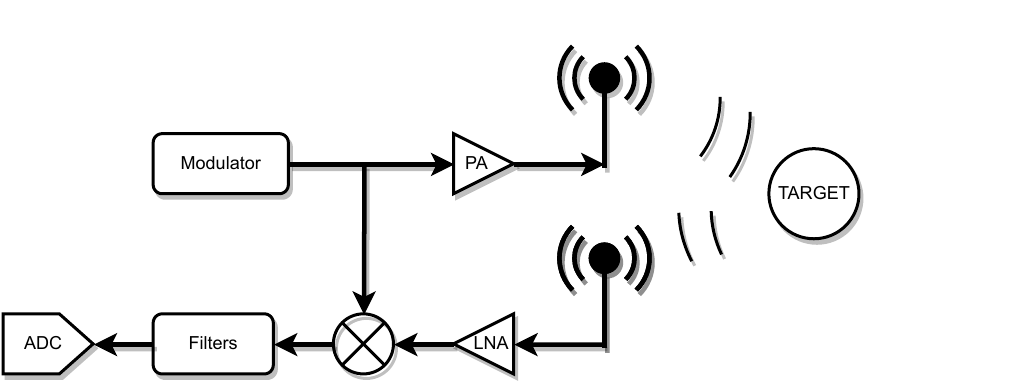}
    \caption{Simplified block diagram of a typical \gls{fmcw} radar system.}
    \label{fig:radar_block_diagram}
\end{figure}

Radar systems emit an electromagnetic wave signal (known as the illumination signal), which eventually hits and is reflected by the target. The reflected (echo) signal contains information about the target, that can be extracted.
The properties of the illumination signal can differ significantly in frequency, modulation, and other characteristics, defining different radar technologies, each suitable for different applications.
\begin{itemize}
    \item Pulse radars transmit high-frequency signals in small bursts and exploit the propagation delay and the antenna placement to extract information about the position of the target.
    \item \gls{cw} radars illuminate the target with continuous power and exploit the Doppler effect to estimate its velocity.
    \item \gls{fmcw} radars add frequency modulation to the illumination signal. This property allows estimating simultaneously the velocity and the distance of the targets.
\end{itemize}
The capability of estimating both the distance and the velocity of the target makes the \gls{fmcw} type a common choice for automotive and industrial applications.
A simplified block diagram of a typical \gls{fmcw} radar system is shown in \cref{fig:radar_block_diagram}.
The modulator is responsible for generating the correct waveform for the illumination signal.
The signal is amplified to the desired power level and transmitted by one or more antennas. At the same time, the echo signal is picked up by the receiving antenna, amplified, and mixed with the transmitted signal.
The mixing process generates a signal with a new frequency and phase equal to the difference between the frequency and phase of the two input signals.
For example, given two inputs signals
\begin{align}
    y_1(t) &= A\exp(2\pi jtf_1 + \phi_1) & y_2(t) &= A\exp(2\pi jtf_2 + \phi_2)
\end{align}
the output of the mixer will be
\begin{equation}
    x(t) = A\exp(2\pi jt(f_2-f_1) + \phi_2-\phi_1)
\end{equation}
This signal is called Intermediate Frequency (IF) signal or beat signal.
It is typically much lower in frequency with respect to the transmitted signal and can be sampled with a traditional Analog to Digital Converter (ADC).

\begin{figure}
    \centering
    \includegraphics[width=\linewidth]{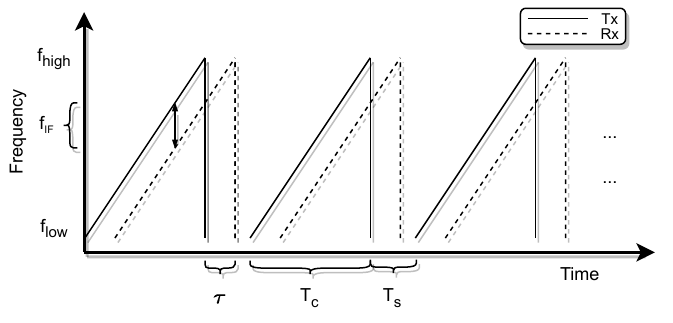}
    \caption{Transmitted and Received signals in \gls{fmcw} radars. Typically $T_c~\ll~T_s$.}
    \label{fig:fmcw_diagram}
\end{figure}

A common modulation for \gls{fmcw} radars is the linear one, where the frequency is linearly increased with time. This signal is also often referred to as \textit{chirp} and can be expressed as
\begin{equation}\label{eq:chirp}
s(t) = A_{tx}\exp{(2\pi jt(f_{low} + St))}
\end{equation}
where $A_tx$ is the signal amplitude, $f_{low}$ is the starting frequency and $S$ is the chirp slope, equivalent to $\frac{f_{high}-f_{low}}{T_c}$ with $T_c$ being the chirp duration. $f_{high}-f_{low}$ is also defined as modulation bandwidth $B$.

\subsection{Distance Measurement}
Given a reflective object at distance $d$, the echo signal is received with a round-trip time delay $\tau = 2d/c$ where $c$ is the speed of light. This effect is depicted in \cref{fig:fmcw_diagram}.
This time delay is taken into account in the mixer, which will generate an IF signal equal to 
\begin{align}\label{eq:if_signal}
y(t) &= A\exp{(2\pi jt(S\tau) + \phi)} & \text{with}& & \tau &=\frac{2d}{c}
\end{align}

\Cref{eq:if_signal} shows that the distance of the targets is proportional to the frequency of the IF signal, whose spectrum shows peaks corresponding to the target range. The spectrum can be evaluated with a \gls{fft}, which is often called \textit{Range \gls{fft}}.

From Fourier transform theory, we know that in a window of duration $T_c$ we can only resolve frequencies larger than $\frac{1}{T_c}$. From the round-trip time, we know that the minimum frequency $f_{min} = 2d_{min}/S$.
From these to equations we can derive that the distance resolution $d_{min}$ only depends on the bandwidth $B$, as shown in \cref{eq:distance_res}.
\begin{align}\label{eq:distance_res}
f_{min} &> \frac{1}{T_c}   &   f_{min} &= \frac{2d_{min}}{c}S  &   d_{min} &> \frac{c}{2ST_c} = \frac{c}{2B}
\end{align}

\subsection{Velocity}
The radial velocity of the target is estimated by observing the phase of two consecutive chirps with a small time spacing $T_s$.
Given a sufficiently small $T_s$, the distance of the target in the range \gls{fft} will be unchanged across the chirps.
However, the phase difference depends on the variation of the round trip time $\Delta \tau$ 
\begin{align}\label{eq:phase}
    \Delta \phi_{IF} &= 2\pi f_{low}\Delta \tau       &       \Delta \phi_{IF} &= \frac{4\pi \Delta d}{\lambda}
\end{align}
From \cref{eq:phase} we can derive the angular velocity caused by the moving target, which is equal to the phase difference across the two chirps.
\begin{align}\label{eq:omega}
    \omega:= \Delta \phi &= \frac{4\pi vT_s}{\lambda}    &   v &= \frac{\lambda\Delta\phi}{4\pi T_s}
\end{align}

The velocity of the target can be resolved without ambiguity when $\Delta \phi < \pi$, which sets the limit for the maximum velocity as $v < \frac{\lambda}{4T_s}$

In the case of multiple targets with different speeds at the same range, the velocity for both targets can be estimated by increasing the number of equi-spaced chirps $N$.
A sequence of $N$ chirps is often referred to as a radar \textit{frame}.
The velocities can be derived with a complex \gls{fft}, called \textit{Doppler \gls{fft}}, phasors corresponding to each range.

From the properties of the \gls{fft}, two frequencies $\omega_1$ and $\omega_2$ can be separated if $|\omega_1-\omega_2|>2\pi/N$. Taking \cref{eq:omega} in consideration we can calculate the velocity resolution $v_{min}$
\begin{align}
    \Delta \omega &= \frac{4\pi vT_s}{\lambda} > \frac{2\pi}{N} & v_{min} := v > \frac{\lambda}{2T_sN}
\end{align}

\subsection{Angle of Arrival}
Radar systems can also estimate the \gls{aoa} of the signal thanks to an array of receiving antennas. Large antenna arrays (with many receivers) can differentiate multiple targets with equal speed and distance, by evaluating a third \gls{fft} on the antenna dimension.
Since the radar evaluated in this paper has a small antenna array (of only 2 receivers on azimuth and elevation), we omit the details of \gls{aoa} evaluation.
However, it is still worth mentioning that the angular resolution $\Theta_{min}$ increases with the length of the array, following the relation
\begin{align}
     \Theta_{min} &= \frac{2}{N}
\end{align}
where $N$ is the number of antennas and d is the antenna spacing, and the antenna spacing is optimal with $d_{spacing}=\lambda/2$.
\section{Evaluation Setup}\label{sec:setup}
This paper evaluates the capability of a novel low-power radar sensor from Infineon Technologies for autonomous racing on our F1TENTH platform.
In particular, we focus on the evaluation of the capabilities for estimating the velocity and position of the opponents placed in front of the car, with the aim of improving the overtaking maneuvers.

The selected sensor device operates in the \SI{60}{\giga\hertz} band and embeds one transmitting and three receiving antennas directly in the package (\gls{aip}), which simplifies the integration in existing systems by removing the need for high-frequency antenna design expertise.
The 60GHz frequency band and the associated bandwidth provide a range resolution of about \SI{3}{\cm}, which is suitable for our racing scenario.
The peak transmission power is \SI{5}{\dBm}, which results in a limited maximum range below \SI{10}{\meter}. However, the reduced form factor and the low-power nature of the device make it suitable for small-scale, battery-operated applications, such as nano drones and small vehicle models, which could exploit the same subsystem to detect other moving or static objects. 

\begin{figure}
    \centering
    \includegraphics[trim={0 1cm 0 1cm},clip,width=\linewidth]{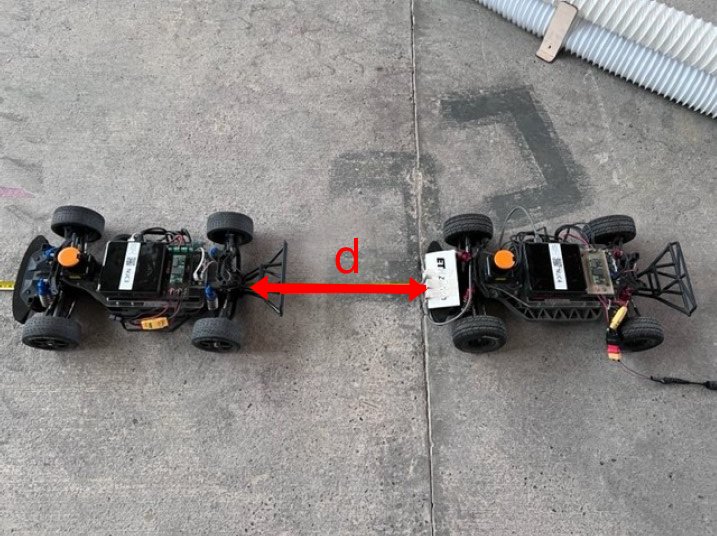}
    \caption{Initial state of the evaluation setup with the initial distance \textit{d} marked in red.}
    \label{fig:two_cars_pic}
\end{figure}

To have an accurate evaluation we concluded two different experiments with a similar setup, which we describe shortly.
Two vehicles were used for the evaluation. The radar subsystem was attached to the front of the first car, facing towards the front of the vehicle, and connected to the Intel NUC via USB.
A custom \gls{ros} driver was developed to interface the USB device, configure it, and retrieve the data.
The second vehicle was placed in front, in the field of view of the radar, and acted as a target for the radar, as seen in \cref{fig:two_cars_pic}.

\subsection{Experimental evaluation}
This first experiment was designed to study the accuracy of the radar in tracking the distance and the velocity of an opponent car in the scenario with the lowest interference.
For this setup, the car equipped with the radar did not drive.
The second car was placed in front of the first one and manually controlled to drive away at different speeds.
The radar data was logged alongside the odometry data of the second car, in order to have a partial ground truth of the velocity readings. It must be noted that the odometry estimation for velocity is less accurate since it depends on indirect measurements.

In the second experiment, both cars were driving at approximately the same speed, both manually controlled.
This experiment served the purpose of observing the effect of the ego-velocity of the car on the radar data, which will be discussed in \cref{sec:results}.

\section{Experimental Results}\label{sec:results}
For this evaluation, the signal processing on the radar signal was kept as simple as possible, only using standard FFT processing.
This evaluation allowed us to properly identify the benefits that the additional radar data could bring, as well as the challenges of extracting the important information depending on the context.

\begin{figure}
    \centering
    \includegraphics[width=\linewidth]{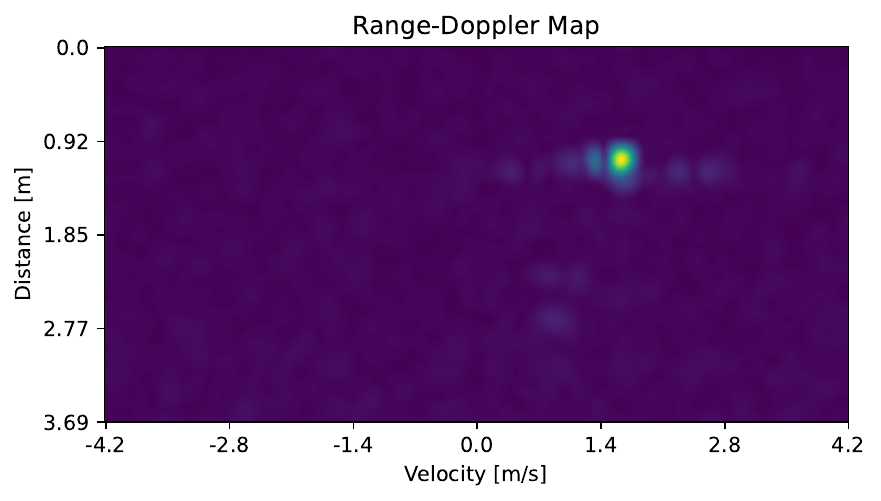}
    \caption{Sample range-doppler map from the static experiments. The moving target can be seen as a high-energy point.}
    \label{fig:doppler_map_static}
\end{figure}

The radar is configured to produce radar frames at \SI{20}{\hertz}.
Each frame is composed by \num{64} chirps, and each chirp is sampled \num{64} times at \SI{2}{\mega\hertz}.
The chirp timing is set in order to allow a maximum range of \SI{3.7}{\meter}, a maximum velocity of \SI{8}{\meter\per\second}, and a velocity resolution of \SI{0.25}{\meter\per\second}.
The current hardware is limited to a bandwidth of about \SI{5}{\mega\bit\per\second} for radar data, which sets an upper bound on the resolution.
DC removal and a Hanning window are used to improve the quality of the resulting map and reduce noise.
The data is zero-padded to \num{256} samples before evaluating the spectra, in order to increase the resolution.
A sample range-doppler map is shown in \cref{fig:doppler_map_static}.

\subsection{Static Measurements}
\begin{figure}
    \centering
    \includegraphics[width=\linewidth]{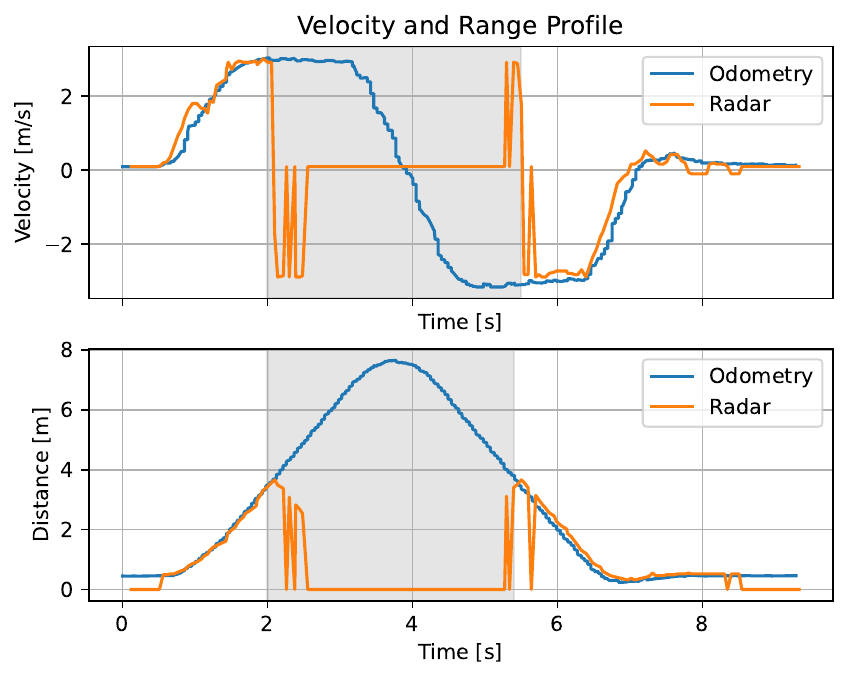}
    \caption{In orange, the speed (top) and distance (bottom) estimated with the radar, in blue the odometry data. Radar data in the grey area is out of range.}
    \label{fig:velocity_range_plot}
\end{figure}

The static measurements are evaluated to characterize the sensor in the best case. They show that distance and velocity can be easily extracted from the range-doppler maps by tracking the max-energy point.
This result shows that, despite the low power output, the radar is capable of detecting moving targets in its range reliably in optimal conditions.
\cref{fig:velocity_range_plot} shows a quantitative view of the range and velocity estimations from the range-doppler map with respect to the values estimated by the opponent car.
The grey area marks the time when the opponent car was outside the maximum range of the radar, and therefore not visible.

We estimated the range and velocity discrepancy for two different top velocities, each with five repeated experiments to reduce the effect of variability.
The errors are evaluated only within the maximum range of the radar, and can be seen in \cref{tab:errors}.

\begin{table}[]
    \centering
    \begin{tabular}{l|c|c|c}
         & $V_{max}$ [\SI{}{\meter\per\s}]  & $R$ RMSE [\SI{}{\meter}] & $V$ RMSE [\SI{}{\meter\per\s}] \\
         \hline \\[-0.8em] 
        Exp. 1 & 3.0 & $0.21 \pm 0.29$ & $0.39 \pm 0.19$ \\
        Exp. 2 & 0.5 & $0.05 \pm 0.03$ & $0.03 \pm 0.01$ \\
    \end{tabular}
    \caption{Evaluation of range and velocity error with respect to the odometry data from the front car. \textit{\textit{R}} indicates range and \textit{V} indicates velocity.}
    \label{tab:errors}
\end{table}

\subsection{Dynamic Measurements}
\begin{figure}
    \centering
    \includegraphics[width=\linewidth]{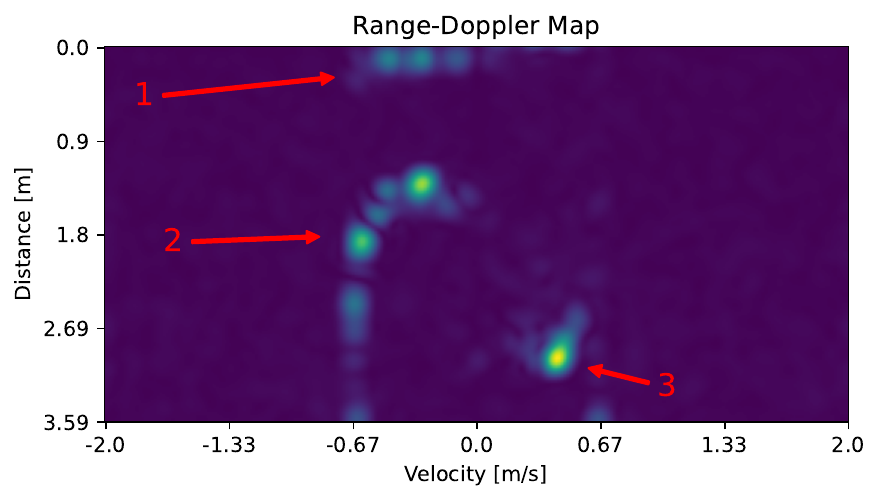}
    \caption{Range-doppler map acquired with both vehicles in motion. Both the opponent (3) and the environment (1,2) are visible in the range-doppler space.}
    \label{fig:doppler_map_dynamic}
    \vspace{-1em}
\end{figure}

In the case of a moving radar, the relative velocity of the environment with respect to the radar is not zero.
This results in multiple targets in the range-doppler map, whose energy depends on properties such as the reflectivity of the material, the angle, and the distance to the radar. In our tests, most of the reflections are caused by the track boundaries. In \cref{fig:doppler_map_dynamic} such targets appear on the side of negative velocity (2), forming the characteristic curved pattern.
This is due to the environment being projected into the map.
Objects in the far field appear at all distances, with a velocity approximately equal to the ego velocity of the car.
However, the radial velocity of the track boundaries lowers as we consider points closer to the car since the relative angle also increases. This can be observed from the curved pattern.
Some reflections from the ground can also be observed at close range (1).

In the frame shown, the opponent vehicle is still visible in the range-doppler map (3), as it is on the positive velocity side.
However, proper tracking in the range-doppler space requires further processing, since the opponent could be masked by the environment reflections at times.
This demonstrates the necessity of a more sophisticated filtering technique that will be addressed in future work. Specifically, in the context of autonomous racing, we believe we can exploit the knowledge of the environment and the odometry information of the car to isolate the opponent in the range-doppler space.
At the same time, we expect that our method will increase the robustness of the LiDAR point cloud by validating LiDAR points against their expected relative velocity, allowing for LiDAR filtering within highly dynamic environments.
Finally, the fusion method will also improve the ability to correctly classify static and dynamic obstacles in racing conditions.

\section{Conclusions and Future Work}\label{sec:conclusion}
A novel low-power \gls{fmcw} radar sensor was evaluated in the context of autonomous racing.
We evaluated the accuracy of distance and velocity tracking in a radar-static scenario, showing that despite the radars low transmission power, in the range of 10s of milliwatts, the sensor is capable of tracking distance and velocity of the target with relatively low tracking error of \SI{0.21}{\metre} and \SI{0.39}{\metre \per \second} respectively.

Dynamic experiments with the radar on a moving car were also conducted, in order to simulate a more realistic racing context.
The subsequent results shed light on the challenges that a dynamic scenario will pose for the velocity estimation of the target, posing a starting point for incremental research on a novel LiDAR-in-radar sensor fusion algorithms.

\section*{Acknowledgment}
 The authors would like to thank Steven Peter, who developed and tested the \gls{ros} driver to interface the radar sensor into the racing stack.

\bibliographystyle{IEEEtran}
\bibliography{refs} 

\end{document}